\documentclass{article} 
\usepackage{iclr2021_conference,times}
\usepackage{booktabs}
\usepackage{graphicx}
\usepackage{amsmath}
\usepackage{multirow}
\usepackage[linesnumbered, ruled, vlined]{algorithm2e}
\usepackage{algorithmic}
\usepackage{mathtools, nccmath}
\usepackage[normalem]{ulem}
\usepackage{float}

\usepackage{hyperref}
\usepackage{url}
\usepackage{color}

\iclrfinalcopy

\title{Towards creativity characterization of generative models via group-based subset scanning}

\author{Celia Cintas \\
IBM Research Africa\\
Nairobi, Kenya\\
\And
Payel Das\\
IBM Research\\
Yorktown Heights, NY, USA\\
\And 
Brian Quanz\\
IBM Research\\
Yorktown Heights, NY, USA\\
\And
Skyler Speakman\\
IBM Research Africa\\
Nairobi, Kenya\\
\And 
Victor Akinwande\\
IBM Research Africa\\
Nairobi, Kenya\\
\And
Pin-Yu Chen\\
IBM Research\\
Yorktown Heights, NY, USA\\

}

\begin{document}

\maketitle
\vspace{-0.65cm}
\begin{abstract}
Deep generative models, such as Variational Autoencoders (VAEs), have been employed widely in computational creativity research. However, such models discourage out-of-distribution generation to avoid spurious sample generation, limiting their creativity. Thus, incorporating research on human creativity into generative deep learning techniques presents an opportunity to make their outputs more compelling and human-like.  As we see the emergence of generative models directed to creativity research, a need for machine learning-based surrogate metrics to characterize creative output from these models is imperative. 
We propose group-based subset scanning to quantify, detect, and characterize creative processes by detecting a subset of anomalous node-activations in the hidden layers of generative models. 
Our experiments on original, typically decoded, and ``creatively decoded''~\citep{das2019toward} image datasets reveal that the proposed subset scores distribution is more useful for detecting creative processes in the activation space rather than the pixel space. Further, we found that creative samples generate larger subsets of anomalies than normal or non-creative samples across datasets. The node activations highlighted during the creative decoding process are different from those responsible for normal sample generation.
\end{abstract}

\vspace{-0.55cm}
\section{Introduction}
\vspace{-0.3cm}
Creativity is a process that provides novel and meaningful ideas~\citep{boden2004creative}. Current deep learning approaches open a new direction enabling the study of creativity from a knowledge acquisition perspective. Novelty generation using powerful deep generative models, such as Variational Autoencoders (VAEs)~\citep{kingma2013auto,rezende2015variational}  and Generative Adversarial Networks (GANs)~\citep{goodfellow2014generative}, have been attempted. 
However, such models discourage out-of-distribution generation to avoid instability and decrease spurious sample generation, limiting their creative generation potential. 
Novelty of the generated samples is often used as a proxy for human perception of creativity in those studies. Therefore, earlier studies mostly focus on estimating the novelty of generated samples, without explicitly considering to the creativity aspect of human perception. Further, those novelty measures do not connect with the generative model features in a quantitative manner, which can provide explanation of the creative generation process. 
The design of creativity evaluation schemes is as essential as developing creative generative methods. Multiple aspects of creativity need to be better defined to allow the research community to develop and test hypotheses systematically ~\citep{cherti2017out}. There are multiple surrogate metrics for novelty~\citep{wang2018generative,ding2014experimental,kliger2018novelty} in the literature; however, the ultimate test of creativity is done by human inspection.  Human labelling has been used to evaluate deep generative models~\citep{dosovitskiy2016learning,lopez2018human} or as a part of the generative pipeline ~\citep{lake2015human,salimans2016improved}. Although human judgement of creativity has numerous drawbacks,  such as annotation is not feasible for large datasets due to its labor-intensive nature, operator fatigue, and intra/inter-observer variations related to subjectivity, it is still crucial to check how humans perceive and judge generated artifacts.
This paper proposes a method designed to detect and characterize when the generative model produces a creative artefact as per a human evaluator. We employ group-based scanning to determine whether a given batch of generated processes contains creative samples using an anomalous pattern detection method called group-based subset scanning~\citep{neill-ltss-2012,mcfowland-fgss-2013}. 

\vspace{-0.3cm}
\section{Proposed Approach: Group-based subset scanning over the creative decoder activation space}\label{sec:groupscan}
\vspace{-0.2cm}
\begin{figure}[t]
    \centering
    \includegraphics[width=0.8\textwidth]{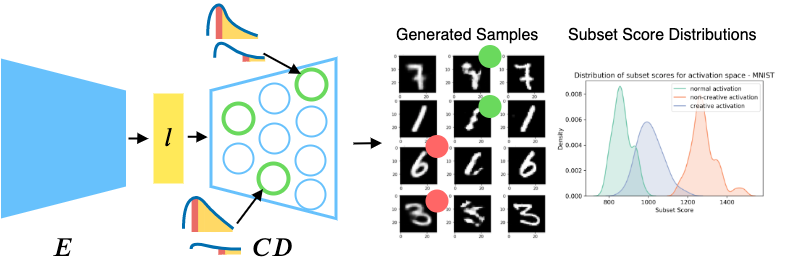}
    \caption{Overview of the proposed approach. First,  we analyze the distribution of the activation space of the Creative Decoder $CD$. After we extracted the activations from the model for a set of latent vectors $l$, we compute the empirical $p-$values followed by the maximization of non-parametric scan statistics (NPSS). Finally, distributions of subset scores for creative, non-creative processes are estimated, a subset of samples and the corresponding anomalous subset of nodes in the network are identified.}
    \label{fig:overview}
\end{figure}
A visual overview of the proposed approach is shown in Figure~\ref{fig:overview}. 
Subset scanning treats the 
creative quantification and characterisation problem as a search for the {\em most anomalous} subset of observations in the data.  This exponentially large search space is efficiently explored by exploiting mathematical properties of our measure of anomalousness. Consider a set of samples from the latent space  $X =\{X_1 \cdots X_M\}$ and nodes $O = \{O_1 \cdots O_J\}$ within the creative decoder $CD$. Where $CD$ is a generative neural network capable of producing creative outputs~\citep{das2019toward}. 
Let $X_S \subseteq X$ and $O_S \subseteq O$,  we then define the subsets $S$ under consideration to be $S = X_S\times O_S$. The goal is to find the most anomalous subset:%
\begin{equation}
    S^{*}=\arg \max _{S} F(S)
\end{equation}
where the score function $F(S)$ defines the anomalousness of a subset of samples from the latent space and node activations. Group-based subset scanning uses an iterative ascent procedure that alternates between two steps: a step identifying the most anomalous subset of samples for a fixed subset of nodes, or a step that identifies the converse.
There are $2^M$ possible subsets of samples, $X_S$, to consider at these steps.  However, the Linear-time Subset Scanning property (LTSS) \citep{neill-ltss-2012,speakman_penalized} reduces this space to only $M$ possible subsets while still guaranteeing that the highest scoring subset will be identified.  This drastic reduction in the search space is the key feature that enables subset scanning to scale to large networks and sets of samples.  

\textbf{Non-parametric Scan Statistics (NPSS)} Group-based subset scanning uses 
NPSS that has been used in other pattern detection methods \citep{mcfowland-fgss-2013,mcfowland-tess-2018,feng-npss_graph-2014,cintas2020detecting,akinwande2020identifying}. Given that NPSS makes minimal assumptions on the underlying distribution of node activations, our approach has the ability to scan across different type of layers and activation functions. 
There are three steps to use non-parametric scan statistics on model's activation data. The first is to form a distribution of  ``expected''  activations at each node ($H_0$).  We generate the distribution by letting the regular decoder process samples that are known to be from the training data (sometimes referred to as ``background'' samples) and record the activations at each node. The second step involves scoring a group of samples in a test set that may contain creative or normal artifacts.  We 
records the activations induced by the group of test samples and compares them to the baseline activations created in the first step.  This comparison results in a $p$-value at each node, for each sample from the latent space in the test set. 
Lastly, we quantify the anomalousness of the resulting $p$-values by finding $X_S$ and $O_S$ that maximize the NPSS, which quantify  how much an observed distribution of $p$-values deviates from the uniform distribution. 

Let $A^{H_0}_{zj}$ be the matrix of activations from $l$ latent vectors from training samples at each of $J$ nodes in a creative decoder layer.  Let $A_{ij}$ be the matrix of activations induced by $M$ latent vectors in the test set, that may or may not be novel.  
Group-based subset scanning computes an empirical $p$-value for each $A_{ij}$, as a measurement for how anomalous the activation value of a potentially novel sample $X_i$ is at node $O_j$. 
This $p$-value $p_{ij}$ is the proportion of activations from the $Z$ background samples, $A^{H_0}_{zj}$, that are larger or equal to the activation from an evaluation sample 
at node $O_j$.
\useshortskip
\begin{equation}
    p_{ij} = \frac{1+\sum_{z=1}^{|Z|} I(A^{H_0}_{zj} \geq A_{ij} )}{|Z|+1}
\end{equation}
Where $I(\cdot)$ is the indicator function. A shift is added to the numerator and denominator so that a test activation that is larger than \emph{all} activations from the background at that node is given a non-zero $p$-value.  Any test activation smaller than or tied with the smallest background acivation at that node is given a $p$-value of 1.0. 

Group-based subset scanning processes the matrix of  $p$-values ($P$) from test samples with a NPSS to identify a submatrix $S =X_S \times O_S$ that maximizes  $F(S)$, as this is the subset with the most statistical evidence for having been affected by an anomalous pattern. 
The general form of the NPSS score function is 
\useshortskip
\begin{equation}
F(S)=\max_{\alpha}F_{\alpha}(S)=\max_{\alpha}\phi(\alpha,N_{\alpha}(S),N(S))
\end{equation}
where $N(S)$ is the number of empirical $p$-values contained in subset $S$ and $N_{\alpha}(S)$ is the number of $p$-values less than (significance  level) $\alpha$ contained in subset $S$. 
It has been shown that for a subset $S$ consisting of $N(S)$ empirical $p$-values, $E\left[N_{\alpha}(S)\right] = N(S)\alpha$ ~\cite{mcfowland-fgss-2013}.
Group-based subset scanning attempts to 
find the subset $S$ that shows the most evidence of an observed significance higher than an expected significance, 
$ N_{\alpha}(S) > N(S)\alpha $, for some significance level $\alpha$.

In this work, we use the Berk-Jones (BJ) test statistic as our scan statistic. BJ test statistic~\citep{berk-bj-1979} is defined as:
\useshortskip
\begin{equation}
    \phi_{BJ}(\alpha,N_\alpha,N) = N*{KL} \left(\frac{N_\alpha}{N},\alpha\right)
\end{equation}
where $KL$ refers to the Kullback-Liebler divergence, $KL(x,y) = x \log \frac{x}{y} + (1-x) \log \frac{1-x}{1-y}$, between the observed and expected proportions of significant $p$-values. We can interpret BJ as the log-likelihood ratio for testing whether the $p$-values are uniformly distributed on $[0,1]$.
\vspace{-0.25cm}
\section{Experimental Setup and Results}
\vspace{-0.25cm}
We hypothesize that creative content leaves a subtle but systematic trace in the activation space that can be identified by looking across multiple creative samples. Further, We assume that not all generative models will have the same throughput of creative samples in a batch. Thus, we need to evaluate our method under different proportions to see if even models that generate a small percentage of creative samples can be detected by our method. 
We test this hypothesis through group-based subset scanning over the activation space that encodes \emph{groups of samples} that may appear anomalous when analyzed together.
We apply our approach to the Creative Decoder and scan the both the pixel/input and activation space~\cite{das2019toward}. We used images from MNIST~\citep{lecun1998gradient} and Fashion-MNIST (FMNIST) \citep{xiao2017fashion} datasets.
We quantify detection \emph{power}, that is the method's ability to distinguish between test sets that contain some proportion of creative samples and test sets containing only normal content, using AUC.


\textbf{Datasets and Creative Labelling}
 For human evaluation, 9 evaluators annotated a pool of 500 samples per dataset (we used agreement amongst $>3$ annotators as consensus), generated from either using the Creative Decoder, and regular decoding. Following~\cite{das2019toward}, we used four labels - `not novel or creative (similar to training data)', `novel but not creative (different from training data but does not seem meaningful or useful)', `creative (different from training data and is meaningful or useful)', and `inconclusive'.  

\textbf{Subset Scanning Setup}
We run individual and group-based scanning on node activations extracted from the Creative Decoder. We tested group-based scanning across several proportions of creative content in a group, ranging from $10\%$ to $50\%$. We used $Z=250$ latent vectors to obtain the background activation distribution ($A^{H_0}$) for experiments with both datasets. For evaluation, each test set had samples were drawn from a set of $100$ normal samples from the regular decoder (separate from Z) and $100$ samples labeled as creative and $100$ non-creative samples(not novel or creative label).

\begin{figure}[t]
    \centering
    \includegraphics[width=0.46\textwidth]{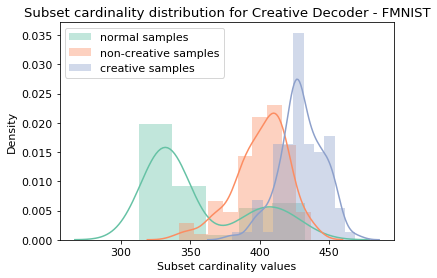}\includegraphics[width=0.45\textwidth]{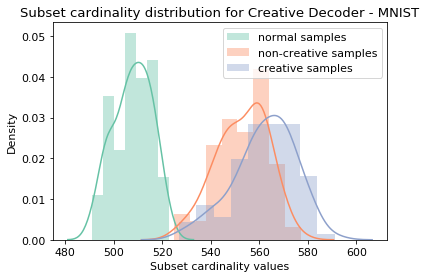}
 \includegraphics[width=0.42\textwidth]{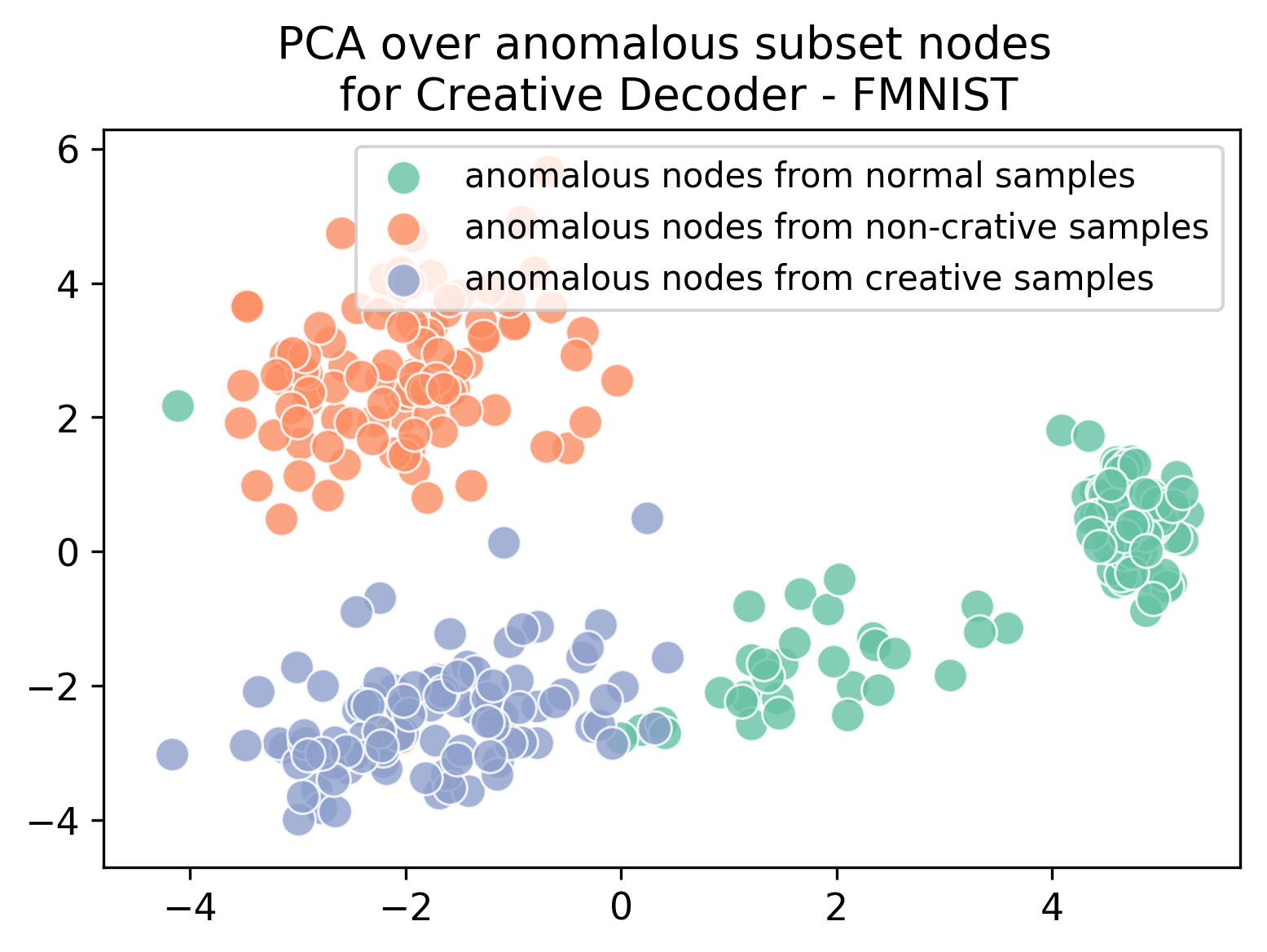}\hspace{0.47cm}\includegraphics[width=0.42\textwidth]{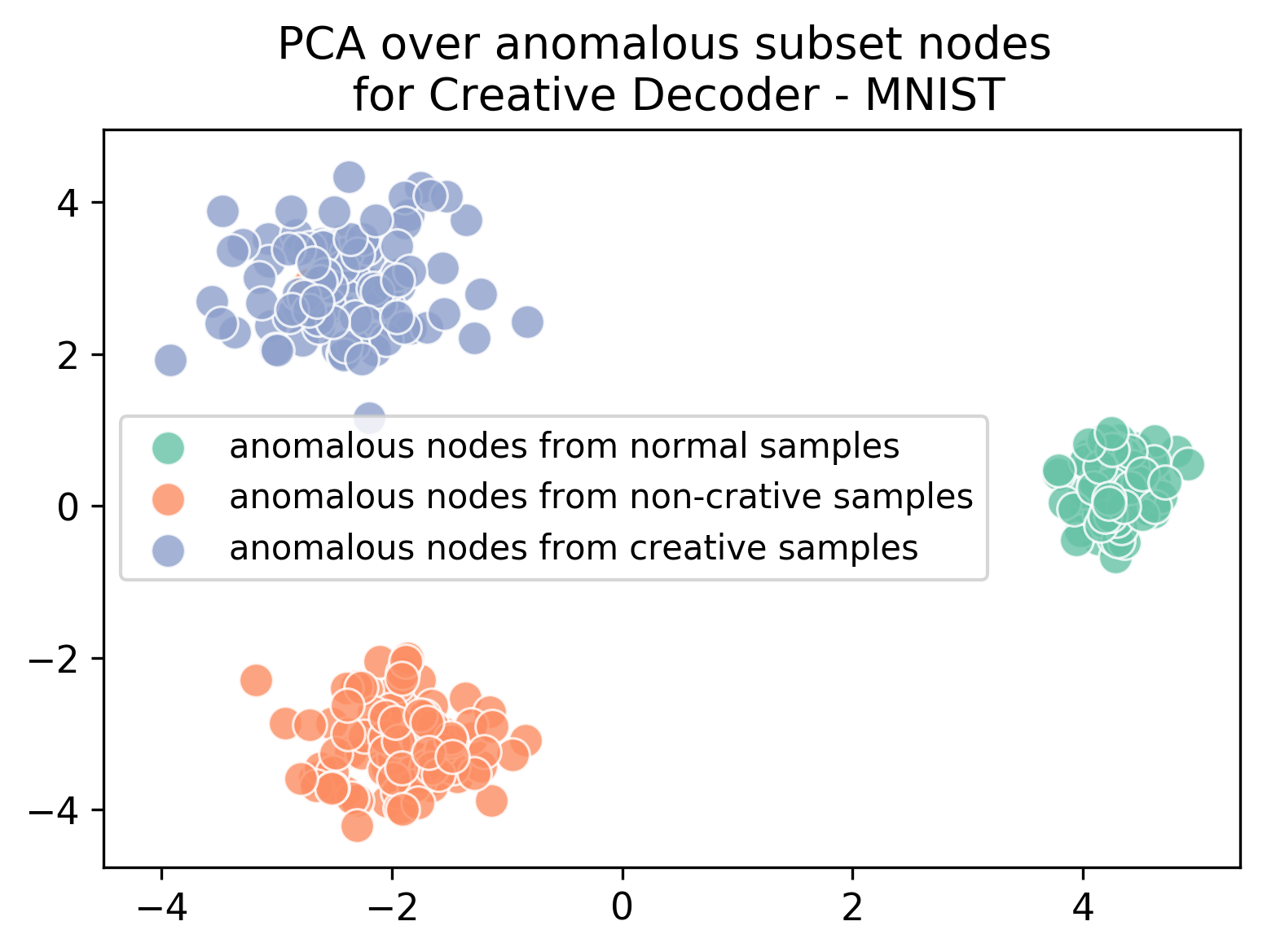}
    \caption{Activations characterization. \textbf{(top)} Subset Cardinality distributions for anomalous subsets for different type of generated samples. \textbf{(bottom)} PCA over anomalous subset nodes for the creative decoder activations under generation of normal, non-creative and creative samples.}
    \label{fig:cardinalitydist}
\end{figure}

\textbf{Results} In Table~\ref{table:ssovercreativedecoder} we present results showing the creative detection capabilities of both activation and pixel spaces. We see that the characterization improves when detecting the creative samples in the activation space, than when we scan over the pixel space. Additionally,  in Figure~\ref{fig:cardinalitydist} for both datasets we  observe a larger extent of anomalous nodes during creative generation compared to normal and non-creative. This observation is consistent with the basic principle of the creative decoding process \citep{das2019toward}. To further inspect the activations, we visualize the principal component projections of the anomalous subset of nodes for  different sets of  samples. As we can see, the  activations for different types of samples are distinctive. Notably, for FMNIST we start noticing some overlap for normal and creative samples. Based on this observation, we hypothesize that as more complex datasets are subject to creative decoding, we will see appearance of more overlapping nodes.

\begin{table}[t]
\centering
 \caption{{\bf Detection Power} (AUC) for group-based and individual subset scanning over pixel and activation space for the Creative Decoder.} 
\begin{tabular}{lcccc}
\toprule
\multicolumn{1}{c}{Space} &
\multicolumn{1}{c}{Dataset} & \multicolumn{3}{c}{Subset Scanning}\\
      \midrule
      & & \multicolumn{1}{c}{50\%} &
       \multicolumn{1}{c}{10\%} &
      \multicolumn{1}{c}{Indv.} \\
 \midrule
Pixel Space & MNIST & $0.971$    & $0.791$             & $0.255$ \\
Activation Space &  MNIST & $\mathbf{0.991}$    & $\mathbf{0.972}$             & $0.531$  \\ 
\midrule
Pixel Space & FMNIST & $0.952$    & $0.743$            & $0.381$  \\ 
Activation Space &  FMNIST & $\mathbf{0.990}$  & $\mathbf{0.962}$           & $0.596$ \\ 
\bottomrule
\end{tabular}

\label{table:ssovercreativedecoder}
\end{table}


\vspace{-0.4cm}
\section{Conclusion and Future Work}
\vspace{-0.3cm}
Our proposed method for creativity detection in machine-generated images works by analyzing the activation space for an off-the-shelf Creative Decoder. We provide both the subset of the input samples identified as creative and the corresponding nodes in the network's activations that identified those samples as creative. In future, we will compare  the proposed creativity quantification with other surrogate metrics for novelty~\citep{wang2018generative,ding2014experimental,kliger2018novelty}. Additionally, we will test across larger datasets and other generative models to understand how we can better capture the human perception of creativity under more complex domains. The final goal is to use this creativity quantification approach  as a control for more efficient generation of artefacts that are consistent with human perception of creativity. 

\bibliography{main}
\bibliographystyle{iclr2021_conference}

\end{document}